\documentclass{article}
\usepackage{spconf,amsmath,graphicx}
\usepackage{times}
\usepackage{soul}
\usepackage{url}
\usepackage[hidelinks]{hyperref}
\usepackage[utf8]{inputenc}
\usepackage[small]{caption}
\usepackage{graphicx}
\usepackage{amsmath}
\usepackage{amssymb}
\usepackage{booktabs}
\usepackage{algorithm}
\usepackage{algorithmic}
\usepackage{multirow}
\usepackage{hyperref}
\usepackage{rotating}
\title{Efficient Scene Text Detection with Textual Attention Tower}
\name{
	L Zhang$^{\star}$\thanks{This work is supported by Ningbo 2025 Key Project of Science and Technology Innovation (2018B10071) and National key research and development plan (2019YFB1311600)}
	YF Liu$^{\star}$
	H Xiao$^{\ddagger}$
	L Yang$^{\star}$
	GM Zhu$^{\star}$
	SA Shah$^{\dagger}$
	M Bennamoun$^{\dagger}$ and
	PY Shen$^{\star}$
}
\address{
	$^{\star}$Xidian University, School of computer science and technology, China\\
	$^{\ddagger}$OrionStar Ltd., China\\
	$^{\dagger}$University of Western Australia, Australia
}

\begin{document}
%
\maketitle
\begin{abstract}
Scene text detection has received attention for years and achieved an impressive performance across various benchmarks.
In this work, we propose an efficient and accurate approach to detect multi-oriented text in scene images.~%
The proposed feature fusion mechanism allows us to use a shallower network to reduce the computational complexity.
A self-attention mechanism is adopted to suppress false positive detections.
Experiments on public benchmarks including ICDAR 2013, ICDAR 2015 and MSRA-TD500 show that our
proposed approach can achieve better or comparable performances with fewer parameters and less computational cost.
\end{abstract}

\section{Introduction}

Oriented scene text detection is one of the most challenging computer vision tasks.
The primary task is to spot text objects in different types of scenes.
Text object may differ in many aspects,
such as font type, texture, orientation.
Moreover,
bounding a single text object with an up-right detection box may lead to a low IoU and detection quality.
Several approaches
\cite{model.east,model.sstd,model.textboxespp}
have already presented impressive successes on various public benchmarks and competitions.

The key of text detection is designing features to distinguish text from backgrounds.
Recently, Convolutional Neural Networks (CNN) based methods such as EAST~\cite{model.east} and IncepText~\cite{model.inceptext}
have achieved the state-of-the-art performance for text detection.
Like other computer vision tasks,
deeper networks provide better performances.
EAST initially adopts PVANET~\cite{model.pvanet} and VGG-16~\cite{model.vgg},
the subsequent approaches used ResNet~\cite{model.resnet} and then ResNeXt~\cite{model.resnext}.

Although several text detection frameworks have been designed, many of the recently proposed models mainly focus on detection precision.
These approaches achieve high precision by complex models and high computational cost, but performance increase is relatively limited.
To address these limitations, we design a novel, computationally efficient and extendable 
network structure to perform competitive detection compared with the former approaches.

\begin{figure}[!tb]
	\centering
	\includegraphics[height=0.6\linewidth]{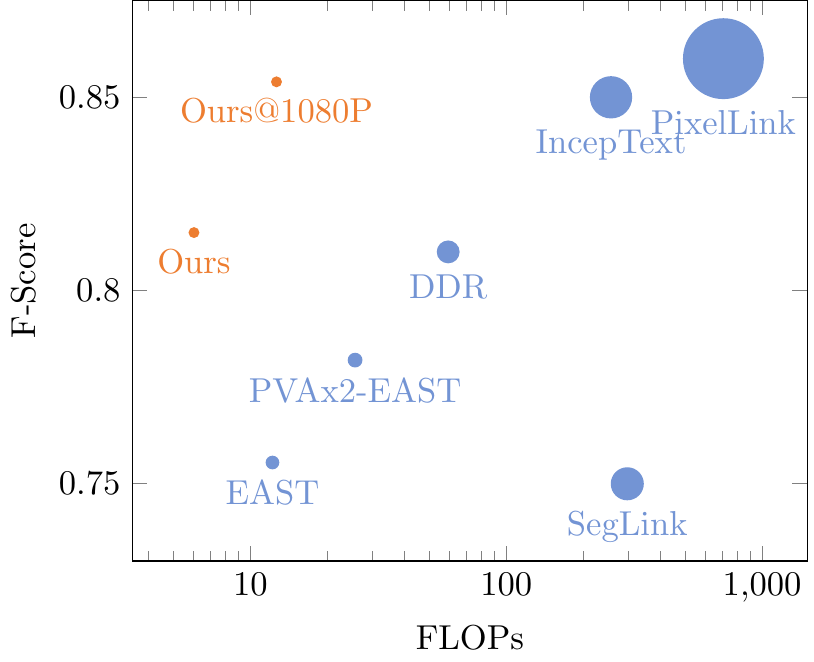}
	\caption{Performance versus floating point operations (FLOPs) on ICDAR 2015 text localization challenge.
		\textbf{Ours} denotes our model evaluated at 720P resolution,
		\textbf{Ours@1080P} denotes our model evaluated at 1080P resolution.
		The area of each node denotes the total parameters in its network.
	}
	\label{fig.compare-plot}
\end{figure}

Our main contribution can be summarized as follows:
\begin{itemize}
	\item We propose a novel, efficient \textit{Textual Attention Tower} (TAT) structure.
	
	\item We evaluate our proposed method on ICDAR 2013, ICDAR 2015 and MSRA TD-500 datasets.
	As shown in Fig. \ref{fig.compare-plot}, the proposed module achieves a significant decrease in the computational cost and a higher accuracy compared with the state-of-the-art models.
\end{itemize}

\section{Proposed Method}

\subsection{Architecture Overview}

\begin{figure*}[ht]
	\centering
	\includegraphics[width=0.6\linewidth]{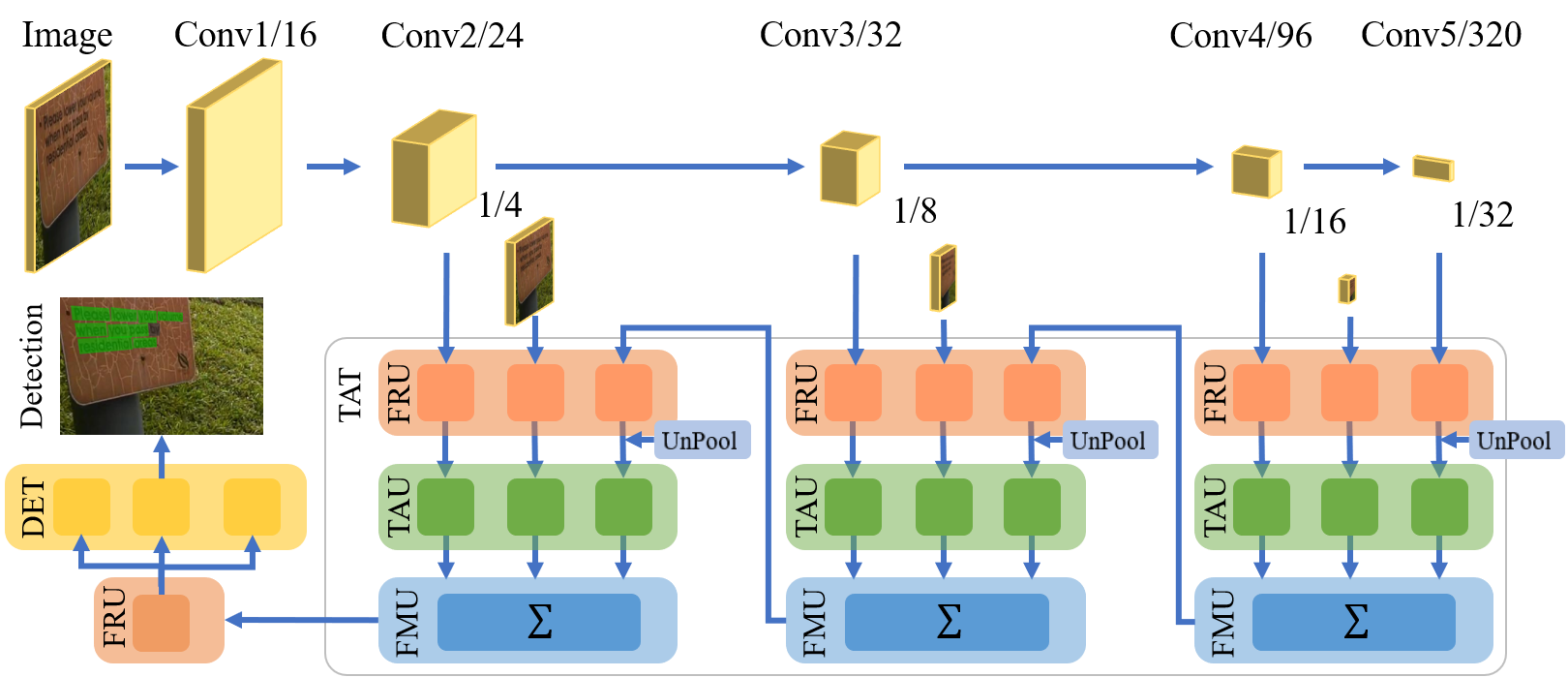}
	\caption{Overview of our proposed model. TAT stands for Textual Attention Tower,
		FRU stands for Feature Refine Unit,
		TAU stands for Textual Attention Unit,
		FMU stands for Feature Mixup Unit,
		DET stands for the Detection Branch.
		Conv i/c stands for the feature maps with $c$ output channels extracted from the \textit{i}-th stage of the MobileNetV2.
		1/n stands for the input image down-sampled to 1/n of the original scale.}
	\label{fig.overall-architecture}
\end{figure*}


An overview of our framework is illustrated in Fig. \ref{fig.overall-architecture}.
The MobileNetV2 architecture is used as the base network. In order to further reduce the computational cost, only the first seven residual blocks of the MobileNetV2 are used.

The key of reducing computational cost and parameter size is our Textual Attention Tower (TAT) architecture,
which 
is designed to fuse the extracted feature maps.
To avoid the degeneration of low level features in deep CNN,
we use the down-sampled input images as extra channels of the intermediate feature maps.

The detection branches in our proposed method are denoted as \textit{DET}.
Inspired by \cite{model.east,model.fots},
we use rotated box (RBOX) to describe text regions.
Thus the \textit{DET} branch is simply $1\times1$ convolutions to map final feature to detections.


\subsection{Textual Attention Tower}
\label{sec.model.tat}

The Textual Attention Tower (TAT) is
designed to 
fuse the feature maps from different stages.
As we adopt segmantation based methodology to regress the geometric information of text regions,
detecting text regions
can be seen as two simple subtasks: text/non-text prediction and distance regression.
Both of these tasks needs large receptive field, and can perform well on tensors with less channels.
As shown in Fig. \ref{fig.overall-architecture}, the TAT has three main parts,
Feature Refine Unit (FRU), Textual Attention Unit (TAU) and Feature Mixup Unit (FMU).

\begin{figure*}
	\begin{center}
		\includegraphics[width=0.6\linewidth]{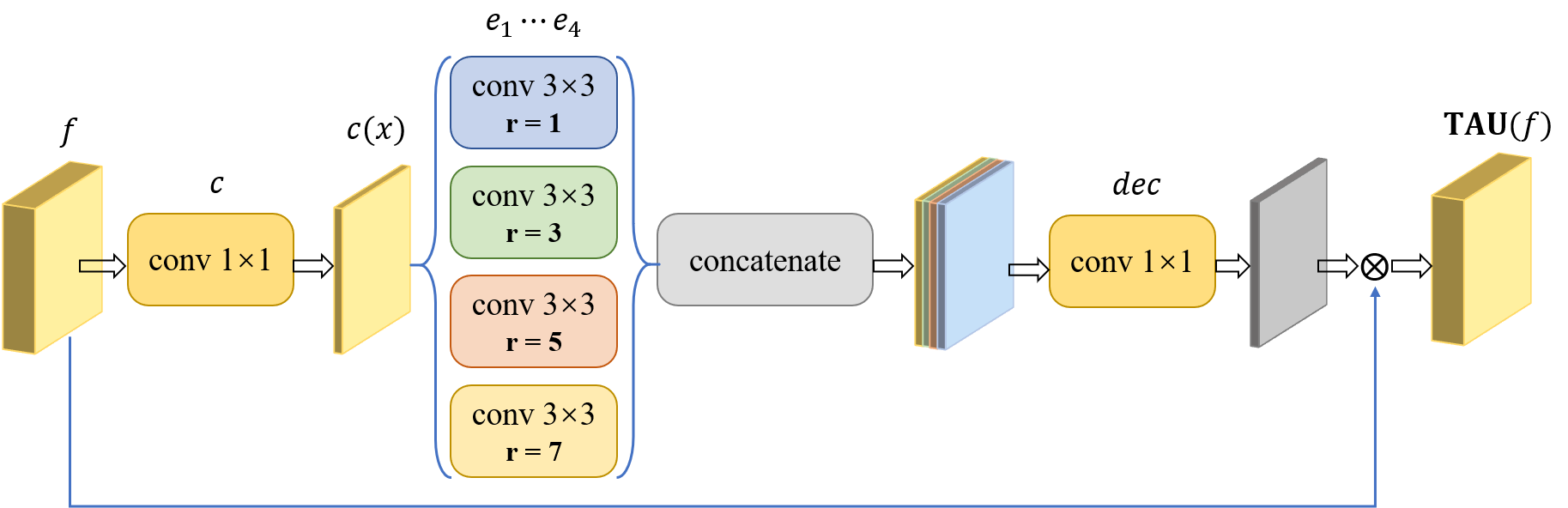}
	\end{center}
	\caption{Detailed structure of our proposed Textual Attention Unit (TAU).}
	\label{fig.tau}
\end{figure*}

\textbf{Feature Refine Unit} The FRU is a ``bottleneck" residual block \cite{model.resnet} to refine the feature maps and reduce its number of channels.
Regardless of the number of input channels, we set the number of output channels of all the FRU modules to 32.
As shown in Fig. \ref{fig.overall-architecture}, we adopt a dedicated FRU module for each input feature map,
and for each down-sampled image, we adopt two cascaded FRU modules to extract the low-level feature maps.

\textbf{Textual Attention Unit} The TAU is a spatial self-attention module designed to 
encode the global context information.
The key idea of TAU is to gather global context information to support the inference at the current position.
We adopt dilated convolution as the basic operation to enlarge the receptive field.
(\textbf{a}) The first part of a TAU module is a standard convolution block $c$ to reduce the number of channels of the input feature map $f$,
which ensures that the subsequent operations can be performed at a low computational cost. (\textbf{b}) The second part is a context encoder, consisting of four dilated convolutional blocks $e_1$, $e_2$, $e_3$ and $e_4$.
Each encoder $e_i$ has an isolated depth-wise convolution layer with a different dilation rate $r = 2i - 1$.
Appropriate padding is configured to ensure that the output of $e_i$ has the same spatial scale as the input feature maps.
The detailed configuration of dilated convolutions used in TAU modules is illustrated on Fig.~\ref{fig.tau}.
(\textbf{c}) The last part of a TAU module is a convolutional decoder block $dec$,
which accepts the concatenated feature maps from all encoders and decodes it to the spatial attention map.
Therefore, the TAU can be formulated as
\begin{equation}
	\mathbf{TAU}(f) = x \otimes \sigma(dec({\langle}e_1(c(x)), \cdots, e_4(c(x)){\rangle}))
\end{equation}
where $\sigma$ is the sigmoid nonlinearity and $\otimes$ is the element-wise multiplication with broadcast semantic.


\textbf{Feature Mixup Unit} The FMU is a simple, element-wise operator to finally fuse all these features.
In all of our experiments, we adopt the element-wise addition as our FMU function.

\subsection{Loss Function}

Our loss function can be formulated as
\begin{equation}
	L = \lambda_{c} L_{c} + \lambda_{d} L_{d} + \lambda_{r} L_{r},
\end{equation}
where $L_{c}$ is the classification loss,
$L_{d}$ is the distance regression loss and $L_{r}$ stands for the rotation regression loss.
$\lambda_{c}$, $\lambda_{d}$ and $\lambda_{r}$ are coefficients to balance the different loss terms.
We set $\lambda_c$ to 1, $\lambda_{d}$ to 2 and $\lambda_{r}$ to 20 in all of our experiments.

To directly maximize the IoU between the candidate and the ground truth,
we adopt dice loss~\cite{loss.dice} as our classification loss,
which can be formulated as
\begin{equation}
	L_{cls} = 1 - 2\frac{\mathbf{S}^{*}\hat{\mathbf{S}}}{\mathbf{S}^{*} + \hat{\mathbf{S}}},
\end{equation}
where $\hat{\mathbf{S}}$
and $\mathbf{S}^{*}$
are the generated and the predicted score maps.

For regression tasks, we adopt IoU loss, which can be formulated as
\begin{equation}
	L_{dis} = -\log\frac{\mathbf{intersection}(\mathbf{D}^{*}, \hat{\mathbf{D}})}{\mathbf{union}(\mathbf{D}^{*}, \hat{\mathbf{D}})},
\end{equation}
where $\hat{\mathbf{D}}$ and $\mathbf{D}^{*}$ are the generated and the regressed distance maps,
$\mathbf{intersection}$ and $\mathbf{union}$ are functions to calculate the area of the intersection and union parts between 
$\hat{\mathbf{D}}$ and $\mathbf{D}^{*}$.

The rotation regression loss is constructed based on the cosine function,
which can be seen as a loose variant of Smoothed-L1 loss.
\begin{equation}
	L_r(\mathbf{R}^{*}, \hat{\mathbf{R}}) = 1 - \cos(\mathbf{R}^{*} - \hat{\mathbf{R}})
\end{equation}


\section{Experiments}


\subsection{Benchmark Datasets and Data Augmentation}

We evaluate our approach on three public benchmark datasets.
All these datasets consist of scene text objects with arbitrary orientations.

\textbf{ICDAR 2013} \cite{DBLP:conf/icdar/2013} consists of 229 training images and 233 testing images of different resolutions,
and most of the text instances are horizontal or near-horizontal.

\textbf{ICDAR 2015} \cite{karatzas2015icdar} is used in the ICDAR2015 Robust Reading Competition Challenge 4.
It contains 1000 images for training and 500 images for testing.
The text bounding boxes
have multi-orientations,
and they are specified by the coordinates of their four corners in a clockwise manner.

\textbf{MSRA-TD500} \cite{yao2012detecting} 
contains 300 training images and 200 testing images.
Different from ICDAR2015, this dataset consists of text lines and separate words.

\textbf{Data Augmentation}.
First, we randomly rotate each image by -15 to 15 degrees.
For each image, we randomly choose a text box as a \emph{kernel},
and then randomly expand the kernel to fit the crop size, which is set to 640 in all of all experiments.
To further enrich the variety of object scales, we perform \emph{subsampling} in the object-centrical cropping processes.
During the expansion of the kernel , we first expand the kernel to $k \times 640$ where $k \sim U(0.5, 2)$,
and then resize the patch to $640\times640$ pixels with a bilinear interpolation.
Color-space jittering and Gaussian blurring are then applied to the cropped image patches.

\subsection{Experimental Setup}

All of our models are trained on a local machine with 4 NVIDIA TITAN X Pascal GPUs.
Our proposed models are 
trained with the ADADELTA~\cite{optim.adadelta} optimizer.
We set the initial learning rate to $1$
and the weight decay coefficient to $1\times10^{-5}$.
The base CNN is initialized with parameters pretrained on ImageNet, the rest of the parameters in our model are initialized according to~\cite{init.he}.
We adopt Cross-GPU Batch Normalization proposed in \cite{model.megdet} to avoid issue that the data distribution have a significant difference with ImageNet

\subsection{Experimental Results}

Our proposed model is trained and evaluated on ICDAR2015, ICDAR2013 and MSRA TD-500.

\textbf{Oriented Text Detection on ICDAR 2015}.
We conduct experiments on ICDAR2015 Challenge 4.
The initialized model is optimized with an ADADELTA optimizer for 600 epochs.
As shown in Table \ref{tab.results-ic15},
when the test images are fed 
at original scale ($1280\times720)$,
our model achieves an F-score of 81.5;
When the test images are upsampled to $1920\times1080$ via bilinear interpolation,
our model reaches 85.4 in F-score without any fine-tuning or model ensembling,
which outperforms most of the previous methods.

\begin{table}[!htb]
	\centering
		\caption{Results on ICDAR2015 Challenge 4 Incidental Scene Text Localization task.
		MS means multi-scale testing, 1080P means test at $1920\times1080$.}
	\begin{tabular}{lrrr}
		\toprule
		Method                                   & Recall & Precision & F-score \\
		\midrule
		SegLink \cite{model.seglink}             &  73.1  &   76.8    & 75.0    \\
		EAST    \cite{model.east}                &  71.4  &   80.6    & 75.7    \\
		EAST MS \cite{model.east}                &  78.3  &   83.3    & 80.7    \\

		PixelLink \cite{model.pixellink}         &  82.0  &   85.5    & 83.7    \\
		IncepText \cite{model.inceptext}         &  80.6&\textbf{90.5}& 85.3    \\
		\midrule
		Ours                                     &  77.8  &   85.8    & 81.5 \\		
		Ours 1080P                               &\textbf{83.2}&87.7&\textbf{85.4}\\
		\bottomrule
	\end{tabular}
	\label{tab.results-ic15}
\end{table}

\textbf{Horizontal Text Detection}.
We fine-tune our model on ICDAR 2013 training set for 200 epochs using the ADADELTA optimizer.
Test images in ICDAR 2013 have different resolution.
We resize all the testing images to $800\times600$ pixels.
As shown in Table.\ref{tab.results-ic13},
our model outperforms most of the existing methods in term of F-score.

\begin{table}[!htb]
	\centering
	\caption{Evaluation results on ICDAR 2013.}
	\begin{tabular}{lrrr}
		\toprule
		Method                                & recall & precision & F-score \\
		\midrule
		TextBoxes++ \cite{model.textboxespp}  &   74.0 &      86.0 &    80.0 \\		
		PixelLink \cite{model.pixellink}      &\textbf{83.6}& 86.4 &    84.5 \\
		SegLink \cite{model.seglink}          &   83.0 & 87.7 & \textbf{85.3}\\
		\midrule
		Ours                                  &   74.3 &      90.3 &    82.1 \\
		Ours MS  & 78.6 &\textbf{92.9} & 85.2 \\		
		\bottomrule
	\end{tabular}
	\label{tab.results-ic13}
\end{table}

\textbf{Long Text Detection}.
The main challenge in MSRA TD-500 is long text detection.
We initialize our model with parameters trained on ICDAR 2015,
and then optimize this model with an ADADELTA optimizer 
in the TD-500 training set.
To fit the testing images with our proposed method,
we down-sample all testing images to $960\times720$ pixels.
As shown in Table \ref{tab.results-td500}, on TD-500,
our proposed method outperforms segmentation based methods,
but cannot surpass the region proposal based frameworks such as IncepText.
The main reason is that without the object-level supervision information,
segmentation based methods usually fail to separate text lines and the surrounding text-like regions.

\begin{table}[!htb]
	\centering
	\caption{Evaluation results on TD-500.}
	\begin{tabular}{lrrr}
		\toprule
		Method                           & Recall & Precision & F-score \\
		\midrule
		DDR \cite{model.ddr}             & 70.0   &    77.0  &  74.0   \\
		EAST \cite{model.east}           & 67.4   &    87.3  &  76.1   \\
		SegLink \cite{model.seglink}     & 70.0   &    86.0  &  77.2   \\
		PixelLink \cite{model.pixellink} & 73.2   &    83.0  &  77.8   \\
		IncepText \cite{model.inceptext} &\textbf{79.0}&\textbf{87.5}&\textbf{83.0}\\
		\midrule
		Ours                             & 75.3   &    81.4  &  78.2   \\		
		\bottomrule
	\end{tabular}
	\label{tab.results-td500}
\end{table}

\textbf{Comparison of Computational Complexity}.
To further compare the computational complexity for our proposed method with the existing methods.
we compare the theoretical FLOPs per pixel of every method over their relative f-score achieved on ICDAR 2015.
Since the multi-scale testing strategies and model-ensembling require significantly more computational resources,
and the increments of FLOPs depend on the implementation,
only performances achieved by a single model without multi-scale testing are included.
As shown in Table \ref{tab.flops-per-pixel},
our proposed models outperform existing models in terms of computational complexity,
and can still achieve a competitive performance.

\begin{table}[!htb]
	\centering
	\caption{Comparison on per pixel computational complexity and corresponding relative F-scores.}
	\begin{tabular}{lrrr}
		\toprule
		Method                           & FLOPs          & F-score \\
		\midrule
		PixelLink~\cite{model.pixellink} & 765.65K        & 83.7    \\ 
		EAST-VGG16~\cite{model.east}     & 310.62K        & 76.4    \\ 
		SegLink~\cite{model.seglink}     & 322.42K        & 75.0    \\ 
		IncepText~\cite{model.inceptext} & 278.53K        & 85.3    \\ 
		DDR~\cite{model.ddr}             &  64.34K        & 81.0    \\ 
		EAST-PVA~\cite{model.east}       &  13.23K        & 75.7    \\ 
		\midrule
		Ours                             &   6.65K        & 81.5    \\ 
		Ours@1080P                       & \textbf{6.65K} & \textbf{85.4} \\ 
		\bottomrule
	\end{tabular}
	\label{tab.flops-per-pixel}
\end{table}

\textbf{Effectiveness of TAT Module}.
Table \ref{tab.results-tat} summarized more detailed results of our models with different settings on ICDAR 2015.
We choose the best-performing model proposed EAST-PVAx2 in \cite{model.east} as baseline in this comparison, which is 
listed in the first line in Table \ref{tab.results-tat}.
When PVAx2 is replaced with full sized MobileNetV2,
the effectiveness of MobileNetV2 itself performs significant performance improvement,
and reduced the computational cost by $25\%$ approximately comparing with EAST.
For the other four model configurations, the last two convolutional blocks in MobileNetV2 are omitted.
With FRUs significantly reduce computational cost,
and TAUs improves the detection precision, our model achieves better performance than EAST with about one quarter FLOPs.

\begin{table}[!htb]
	\newcommand{\Y}{\checkmark}
	\centering
	\small
		\caption{Effectiveness of TAT on ICDAR2015 incidental scene text location task.
		``M'' means ``MobileNetV2'' and ``I'' means use raw input as extra feature.
		``P'', ``R'', ``F'' represent ``Precision'', ``Recall'', ``F-measure'' respectively.}
	\begin{tabular}{ccccccccc}
		\toprule
		M   & FRU & TAU & I   &    R &    P &    F & FLOPs  \\
		\midrule
		&     &     &     & 73.5 & 83.6 & 78.2 & 23.85G \\ 
		\Y   &     &     &     & 73.7 & \textbf{87.8} & 80.1 & 17.75G \\
		\Y   & \Y  &     &     & 77.4 & 83.6 & 80.4 &  \textbf{5.79G} \\ 
		\Y   & \Y  & \Y  &     & 77.2 & 85.8 & 81.3 &  5.85G \\ 
		\Y   & \Y  & \Y  & \Y  & \textbf{77.8} & 85.8 & \textbf{81.5} &  6.03G \\ 
		\bottomrule
	\end{tabular}
	\label{tab.results-tat}
\end{table}

\section{Conclusion}

In this paper, we propose a novel and efficient multi-oriented text detection method from natural scene images.
The main idea of our design is the use of dilated convolution to keep a reasonable yet abundant information for different levels of receptive fields.
Another improvement comes from using homogenous bottlenecks with the base network to refine feature maps.
We achieve a better performance of our proposed technique on three public scene text benchmarks.

\bibliographystyle{IEEEbib}
\bibliography{EFFICIENT_SCENE_TEXT_DETECTION_WITH_TEXTUAL_ATTENTION_TOWER}

\end{document}